\pdfoutput=1
\documentclass[letterpaper, 10 pt, conference]{ieeeconf}
\IEEEoverridecommandlockouts 
\usepackage{amsmath}
\usepackage{balance}
\usepackage{amsfonts}
\usepackage{amssymb}
\usepackage{color}
\usepackage{subfigure}
\usepackage{wrapfig}
\usepackage{authblk}
\usepackage{stfloats}

\newtheorem{definition}{Definition}

\newtheorem{lemma}{Lemma}

\newtheorem{assumption}{Assumption}
\usepackage{microtype}
\usepackage{hyperref}
\usepackage{multirow}
\usepackage[percent]{overpic}
\usepackage[
  subtle
]{savetrees}

\title{\LARGE \bf
Safe Learning-Based Feedback Linearization Tracking Control for Nonlinear System with Event-Triggered Model Update}

\author{
    Zhixuan Wu$^1$,
    Rui Yang$^1$,
    Lei Zheng$^2$, and
    Hui Cheng$^{1*}$
\thanks{$^1$Zhixuan Wu, Rui Yang and Hui Cheng are with the School of Computer Science and Engineering, Sun Yat-sen University, Guangzhou 510006, China.% Corresponding author: Hui Cheng.
        }
\thanks{$^2$Lei Zheng is with the School of Electronics and Information Technology, Sun Yat-sen University, Guangzhou 510006, China
        }
\thanks{$^*$Corresponding author: {\tt\footnotesize chengh9@mail.sysu.edu.cn}}
}

\begin{document}
\maketitle

\begin{abstract}
Learning-based methods are powerful in handling complex scenarios. However, it is still challenging to use learning-based methods under uncertain environments while stability, safety, and real-time performance of the system are desired to guarantee. In this paper, we propose a learning-based tracking control scheme based on a feedback linearization controller in which uncertain disturbances are approximated online using Gaussian Processes (GPs). Using the predicted distribution of disturbances given by GPs, a Control Lyapunov Function (CLF) and Control Barrier Function (CBF) based Quadratic Program is applied, with which probabilistic stability and safety are guaranteed. In addition, the trajectory is optimized first by Model Predictive Control (MPC) based on the linearized dynamics systems to further reduce the tracking error. We also design an event trigger for GPs updates to improve efficiency while stability and safety of the system are still guaranteed. The effectiveness of the proposed tracking control strategy is illustrated in numerical simulations.
\end{abstract}

% \begin{IEEEkeywords}
% Control Architectures and Programming, Machine Learning for Robot Control, Adaptive Control.
% \end{IEEEkeywords}

%===============================================================================
\section{Introduction}

\subsection{Motivation}
Autonomous mobile robots are widely applied in many fields to solve complex tasks.
Robotic systems have reached much success, including autonomous vehicles, mobile robotic platforms for planetary exploration, robotics arms for industrial assembly, and assistance of surgery \cite{Thrum2005probabilistic}.
In each of these scenarios, robots are required to follow a trajectory accurately in order to complete the tasks.
Mobile robots such as quadrotors and self-driving cars may cause severe accidents if they are unable to track the desired trajectory accurately.
Non-mobile robots like industrial assembly robotics arms will at least fail to complete the assembly with an inaccurate tracking process.
Therefore, precise tracking performance is a basic requirement for different robotics systems.

Besides, safety is crucial for dynamic control systems.
Violating safety constraints will cause damages not only to robots themselves but also to humans in many scenarios.
With inaccurate tracking caused by uncertain disturbances, robots will deviate from the desired trajectory and even collide with obstacles. Hence, safe tracking control is required to ensure precise trajectory tracking for a robotic system under uncertain disturbances.

In particular, some uncertain disturbances in the real world are highly dynamic and unpredictable.
These uncertainties make it difficult for the traditional model-based controller to achieve a satisfactory control performance.
Therefore, the desired trajectory controller should adapt to uncertain disturbances online to ensure a high control accuracy and guarantee safety.

\begin{figure}[t]
\centering
\includegraphics[scale=0.4]{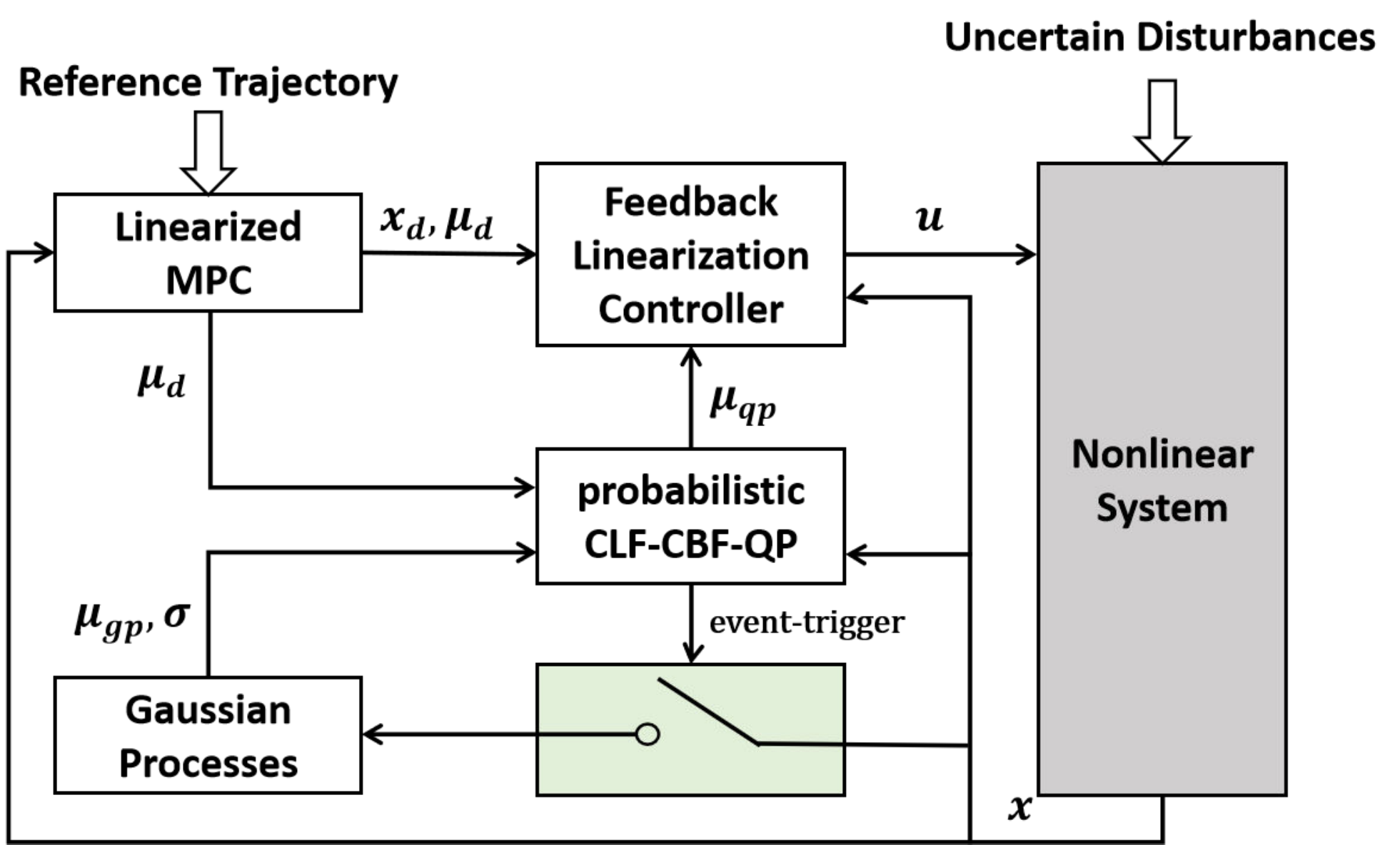}
\caption{Architecture diagram of the proposed control scheme for a nonlinear system under uncertain disturbances. The reference trajectory is first optimized in MPC based on the linearized model. The GPs are used to estimate the disturbances online. The outputs of MPC $\mu_d$ coupled with the outputs of GPs, $\mu_{gp}$ and $\sigma$, are used to obtain $\mu_{qp}$ in CLF-CBF-QP to guarantee stability and safety of the system under high probability. The GPs will be updated in the proper situation determined by an event-triggered update scheme based on the result of CLF-CBF-QP.}
\label{fig:1}
\end{figure}

Motivated by the challenges mentioned above, we propose a novel trajectory tracking control scheme as shown in Fig. \ref{fig:1}.
The proposed method is based on a feedback linearization controller transforming the complex nonlinear system into a linearized system.
Given a reference trajectory for specific tracking tasks, an efficient model predictive control (MPC) combined with the linearized dynamics is deployed to provide the optimized trajectory and control, and to improve the tracking performance utilizing its predictive capability.
The uncertain disturbances are modeled and compensated by Gaussian Processes (GPs) for accurate tracking control under uncertainties.
Considering uncertainties and fidelity of the GP model, we use a probabilistic Control Lyapunov Function- and Control Barrier Function-based Quadratic Program (CLF-CBF-QP), extended from our previous work \cite{Yang2021learning}, to guarantee stability and safety under high probability.
Besides, we design an event-triggered model update scheme for real-time performance and satisfactory model fidelity.
The controller which we call CLBFET (probabilistic CLF-CBF-QP with Event-Trigger for learning-based feedback linearization controller) is proposed to reach safe and accurate tracking under uncertainties for trajectory tracking tasks.

\subsection{Related Work}
% \vspace{-1mm}
To perform accurate tracking, MPC is widely used for complex dynamics systems to anticipate future events and take control actions accordingly.
MPC is combined with feedforward linearization control in \cite{Greeff2018flatness} for quadrotors' trajectory tracking tasks, but the dynamic model errors are ignored which may accumulate and affect the control performance.
MPC based on a hybrid model is proposed in \cite{Hewing2020cautious} and \cite{Torrente2021data}, where model errors are evaluated by GPs and the GP approximation is propagated forward in time.
Stability and safety are considered in learning-based MPC (LBMPC) \cite{Aswani2013provably} for linear systems.
\cite{Capotondi2019an} considers a feedback linearization controller modified by GPs, and constructs MPC with state and control constraints based on it.
The methods \cite{Aswani2013provably}\cite{Capotondi2019an} combine MPC with GPs and consider constraints in MPC, which brings more computational burdens.
In contrast, in this paper, the MPC is based on the linearized system without stability and safety constraints, which is more efficient to maintain predictive proactivity.
The outputs of MPC are then adjusted according to uncertainties and other constraints.

In order to handle stability and safety constraints in a certain environment in real-time, CLF- and CBF-based control methods have been presented for safety-critical systems \cite{Ames2014control}.
Considering environmental uncertainties, a robust CLF-CBF-QP \cite{Nguyen2016optimal} is proposed to consider stability and safety constraints under uncertainties.
It assumes that the environmental uncertainties are bounded and considers the strictest case.
Such a strategy keeps the constraints well but performs relatively poorly in trajectory tracking as its conservative nature of the bounded uncertainties.
In \cite{Choi2020reinforcement}, a Reinforcement Learning (RL) framework estimating uncertainties in CLF and CBF (RL-CLF-CBF-QP) is proposed and numerically validated on a bipedal robot.
However, RL-based methods\cite{Choi2020reinforcement}\cite{Chebotar2017combining}\cite{Castaneda2020improving}, including model-based methods and model-free methods, lack insightful analysis of the learned models or policies.

By contrast, Bayesian model-based methods are used to provide theoretical guarantee for stability and safety analysis under uncertainties.
\cite{Beckers2017stable} guarantees a globally bounded tracking error using CLF with confidence bounds of GP inference without consideration of control constraints.
The confidence of GPs is maximized to stabilize the system under control constraints in \cite{Umlauft2018uncertainty}.
In contrast, to perform the tracking task, we use a fixed high confidence to prevent the tracking performance from being affected by maximizing GP confidence.
In \cite{Zheng2020learning}, the model errors estimated by GPs are used in CLF and CBF to ensure stability and safety instead of achieving accurate tracking.
In this paper, the model errors are compensated by the estimation of GPs to mitigate the tracking performance degradation caused by model uncertainties.
Besides, the methods \cite{Beckers2017stable}\cite{Umlauft2018uncertainty}\cite{Zheng2020learning} may fail to online update GP due to its high computational complexity, which will affect the real-time performance of the algorithms.
In \cite{Castaneda2021gaussian}, the errors in CLF instead of the system dynamics are modeled by pre-trained GPs.
It allows uncertainties in control inputs to be considered and scales well with dimension, but the GP model cannot be reused in other constraints such as safety constraints.
In \cite{Fan2020bayesian}, Bayesian neural networks instead of GPs are used to learn model uncertainties as they found GPs to be computationally intractable with too much training data, although GPs exhibit good performance.
While in our method, the GPs with limited data still perform well as a result of the use of confidence bounds in constraints and a satisfactory model fidelity maintained by the event-triggered update scheme.
An event-triggered scheme is also applied in \cite{Umlauft2020feedback} with an invariant threshold.
We have extended the event-triggered scheme to keep the stability constraints in our probabilistic CLF-CBF-QP.

\subsection{Contributions}
Our main contributions are summarized as follows.
\begin{enumerate}
    \item An online learning-based adaptive tracking control framework is proposed for the trajectory tracking control under uncertain disturbances, where stability and safety are guaranteed under high probability.
    \item An MPC scheme based on linearized dynamics is presented to optimize the trajectory for further improving the tracking performance.
    \item An event-triggered model update scheme for online learning is devised to release computational burden while maintaining the fidelity of GP models. 
    \item The proposed control strategy is validated on the tracking task under uncertain disturbances via simulations.
\end{enumerate}

The remainder of this paper is structured as follows: After formulating the problem in Sec. \ref{sec:problem}, Sec. \ref{sec:methodology} presents the proposed methodology. Numerical simulations of the proposed approach on a quadrotor are shown in Sec. \ref{sec:simulation} followed by a conclusion in Sec. \ref{sec:conclusions}.

%===============================================================================
\section{Problem Statement}
\label{sec:problem}
Consider a nonlinear control affine system with dynamics
\begin{align}
\dot{x}_1=x_2, \quad  \dot{x}_2=f(x) + g(x)u,
\label{eq:dynamics}
\end{align}
with state $x=[x_1,x_2]^\intercal \in X \subset \mathbb{R}^{2n}$, $x_1,x_2\in\mathbb{R}^n$, the state space $X$ is compact, and the controls $u\in\mathbb{R}^n$.
A wide range of nonlinear control-affine systems in robotics such as quadrotors and car-like vehicles can be transformed into this form.
Our analysis is restricted to systems of this form while our results can be extended to systems of higher relative degree \cite{Khalil2002nonlinear} \cite{Nguyen2016exponential}.
In general, on a real system, $f$ and $g$ may not be known exactly.
As a result, we make the following assumptions to make our analysis more tractable.

\begin{assumption}
    The function $f:X\rightarrow \mathbb{R}^n$ is unknown but has a bounded reproducing kernel Hilbert space (RKHS) norm under a known kernel $k$. The function $g:X\rightarrow \mathbb{R}^{n \times n}$ is known and invertible.
\end{assumption}

Based on the assumption, the goal is to design a control strategy that satisfies the following objectives:
\begin{enumerate}
    \item The control strategy adaptively learns and compensates for uncertain disturbances online. 
    \item Asymptotic stability is guaranteed to achieve a high trajectory tracking performance.
    \item Safety is guaranteed.
    \item The control strategy is able to drive the system back to the reference target after deviating from the trajectory as a result of disturbances.
\end{enumerate}

%===============================================================================
\section{Methodology}
\label{sec:methodology}

To achieve the desired goals, we construct the control scheme named CLBFET (probabilistic CLF-CBF-QP with Event-Trigger for learning-based feedback linearization controller). {Given a predefined reference trajectory for specific tasks}, the linearized MPC gives out optimized trajectory and control, which improves the tracking performance. Adaptive control is realized by GPs which is used to estimate and compensate for the uncertain disturbances. Taking the outputs of GPs into account, the probabilistic CLF-CBF-QP solves a QP to guarantee asymptotic stability and safety of the system. It also provides an event trigger to determine whether the GPs need to be updated. We first introduce the feedback linearization controller in Section \ref{sec:fblc}. {Then GPs, probabilistic CLF-CBF-QP and linearized MPC are introduced} in Section \ref{sec:gp}, \ref{sec:CLF-CBF-QP} and \ref{sec:mpc} respectively. Event-triggered update scheme is introduced in Section \ref{sec:et}.

\subsection{Feedback Linearization Control Law}
\label{sec:fblc}
For dynamics (\ref{eq:dynamics}), let $\hat{f}(x)$ be a given nominal model of $f(x)$. We formulate the feedback linearization control law $u$ with pseudo-control component $\mu \in \mathbb{R}^n$:
\begin{equation}
    u = g^{-1}(x)(\mu - \hat{f}(x))
\end{equation}
which brings an approximately linear integrator model
\begin{equation}
    \
    \dot{x}_1 = x_2, \quad \dot{x}_2 = \mu + \delta(x)
\end{equation}
where $\delta(x) = f(x) - \hat{f}(x) \in \mathbb{R}^n$ is the modeling error.

Here we design the pseudo-control component $\mu$ made up of four separate terms. Suppose the optimized reference state $x_{d} = [x_{1d}, x_{2d}]^\intercal \in \mathbb{R}^{2n}$ and control $\mu_{d} \in \mathbb{R}^n$ are given by model predictive control, we have:
\begin{equation}
    \label{eq:4termscontrol}
    \mu=\mu_{d} + \mu_{pd} + \mu_{qp} -\mu_{gp}
\end{equation}
where $\mu_{pd} = K_P(x_{1d} - x_1) + K_D(x_{2d} - x_2)$ is given by a PD controller \cite{Spong2006robot} with the proportional matrix $K_P \in \mathbb{R}^{n \times n}$ and the derivative matrix $K_D \in \mathbb{R}^{n \times n}$, $\mu_{qp}$ is the solution of a quadratic program to guarantee asymptotic stability and safety of the system, and $\mu_{gp}$ is given by {GPs} to compensate the disturbance $\delta(x)$.
{The tracking error is defined as $e = x - x_{d}$ and} $\mu_{pd}$ can be written as:
\begin{equation}
    \label{eq:qpcontrol}
    \mu_{pd} = \begin{bmatrix}-K_P & -K_D\end{bmatrix}e.
\end{equation}
Then we can write the dynamics of the tracking error $e$ as:
\begin{align}
    \label{eq:errdyn1}
    \dot{e}=\begin{bmatrix}\dot{e}_1 \\ \dot{e}_2\end{bmatrix}
    &=\begin{bmatrix} 0 & I \\ -K_P & -K_D \end{bmatrix}e
    +\begin{bmatrix} 0 \\ I \end{bmatrix}(\mu_{qp} -\mu_{gp} + \delta(x)) \\
    \label{eq:errdyn2}
    &=Ae+B(\mu_{qp} -\mu_{gp} + \delta(x)).
\end{align}

\subsection{Gaussian Process Regression}
\label{sec:gp}
As mentioned above, $\mu_{gp}$ is designed to approximate and compensate the disturbance ${\delta}(x)$ using Gaussian Processes (GPs).
A Gaussian Process, one of the stochastic processes, is a generalization of the Gaussian probability distribution \cite{Rasmussen2006GPforML} and has been widely used as a data-driven machine learning model.
As a stochastic process, Gaussian Process is considered as a distribution over functions, and any subset of the inputs obeys a joint Gaussian distribution.
Let $\hat{\delta}_i(x)$ be the approximation of $\delta_i$, it is denoted by
\begin{equation}
\label{eq:gp1}
    \hat{\delta}_i(\mathbf{x})\sim\mathcal{GP}(m_i(\mathbf{x}),k_i(\mathbf{x},\mathbf{x}'))
\end{equation}
where Gaussian process is characterized by a mean function $m(\mathbf{x}): X\rightarrow \mathbb{R}$ and a covariance function $k(\mathbf{x},\mathbf{x}'): X\times X\rightarrow \mathbb{R}$.
{It is a common practice to set the mean function} $m_i(\mathbf{x})=0$, for all $i=1,\cdots,n$ if no prior knowledge is available. The covariance function, also called kernel function, maps two inputs to a scalar output and is specified by a positive-definite kernel.
The squared-exponential (SE) kernel is a widely used kernel, which is denoted by
\begin{equation}
    \label{eq:gpkernel}
    k_i(\mathbf{x},\mathbf{x}')=\sigma_f^2exp\left(-\dfrac{1}{2}(\mathbf{x}-\mathbf{x}')^Tl^{-2}(\mathbf{x}-\mathbf{x}')\right)
\end{equation}
where $\sigma_f$ and $l$ are hyperparameters which describe the prior variance and the length scale respectively. Since (\ref{eq:gp1}) represents only scalar outputs, $n$ independent Gaussian Processes are used to model nonlinear function $\delta:X \rightarrow \mathbb{R}$.
{The calculation time increases with dimensions, and is mainly from the update of the models.
This problem can be mitigated by the event-triggered model updating scheme introduced in Section \ref{sec:et}.
Besides, since each GP is relatively independent, updating each GP in parallel also helps.}
\iffalse
\begin{equation}
    \hat{\delta}(\mathbf{x})=
    \begin{cases}
        \hat{\delta}_1(\mathbf{x})\sim\mathcal{GP}(m_1(\mathbf{x}),k_1(\mathbf{x},\mathbf{x}')),\\
        \vdots \\
        \hat{\delta}_n(\mathbf{x})\sim\mathcal{GP}(m_n(\mathbf{x}),k_n(\mathbf{x},\mathbf{x}')),
    \end{cases}
\end{equation}
\fi
\begin{assumption}
    The state $x$ and the function value ${\delta}(x)$ can be measured with noise over a finite time horizon to make up a training set with N data pairs
    \begin{equation}
        \label{eq:gpdataset}
		\mathcal{D}=\left\{\left(\mathbf{x}^{\left(i\right)},\ y^{\left(i\right)}\right)\right\}_{i=1}^N,\ y^{(i)}={\delta}\left(\mathbf{x}^{(i)}\right)+w_i
	\end{equation}
	where $w_i$ are i.i.d. noises $w_i\sim N\left(0,\sigma_{noise}^2I_n\right)$, $\sigma_{noise}^2\in\mathbb{R}$.
\end{assumption}
Given the dataset $\mathcal{D}$, the Gaussian Process is employed for regression. For any query input $\mathbf{x}^*$, the $j$-th component of the inferred output $y^*$ is jointly Gaussian distributed with the training dataset
\begin{equation}
\begin{bmatrix}
    y_j^* \\
    \mathbf{y}_j
\end{bmatrix}
\sim \mathcal{N}\left(
\begin{bmatrix}
    0 \\
    0
\end{bmatrix},
\begin{bmatrix}
    k_j^*   & \mathbf{k}_j^T \\
    \mathbf{k}_j    & \mathbf{K}_j + \sigma_{noise}^2I
\end{bmatrix}
\right)
\end{equation}
where $k_j^*=k_j(x^*,x^*)\in\mathbb{R}$, $\mathbf{y}_j=[y_j^{(1)},\cdots,y_j^{(N)}]^T\in\mathbb{R}^N$, $\mathbf{k}_j=[k_j(x^*,x^{(1)}),\cdots,k_j(x^*,x^{(N)}]^T\in\mathbb{R}^N$, and
\begin{equation}
\label{eq:gpK}
\mathbf{K}_j=
\begin{bmatrix}
    k_j(x^{(1)},x^{(1)})    &   \cdots  &   k_j(x^{(1)},x^{(N)})    \\
    \vdots  &   \ddots  &   \vdots  \\
    k_j(x^{(N)},x^{(1)})    &   \cdots  &   k_j(x^{(N)},x^{(N)})
\end{bmatrix}\in\mathbb{R}^{N \times N}.    \notag
\end{equation}
It yields
\begin{equation}
	\mu_j(\mathbf{x}^*)=\mathbf{k}_{j}^{T}(\mathbf{K}_j + \sigma_{noise}^{2}\mathbf{I}_N)^{-1}y_{j},
\end{equation}
\begin{equation}
    \label{eq:gpvar}
	\sigma_j^{2}(\mathbf{x}^*)=k_j^*-\mathbf{k}_j^T(\mathbf{K}_j+\sigma_{noise}^2\mathbf{I}_N)^{-1}\mathbf{k_j}.
\end{equation}

\begin{lemma}
    \label{lemma:gp1}
    \cite{Umlauft2018uncertainty}
    For any compact set $X \subset \mathbb{R}^{2n}$ and probability $\varsigma \in (0,1)$ holds
    \begin{equation}
        Pr\{\Vert \mu(x) - \delta(x) \Vert \leq \Vert \beta \Vert \Vert \sigma(x) \Vert, \forall x \in X\} \geq (1-\varsigma)^{2n}
    \end{equation}
    {where $Pr$ denotes probability, and $\beta = [\beta_1, \cdots, \beta_n]^\intercal$ with
    \begin{align}
    \beta_j = \left(2\Vert\delta_j\Vert^2_{k_j}+300\gamma_jln^3\left(\dfrac{N+1}{\varsigma}\right)\right)^{\dfrac{1}{2}}, \quad j=1,\cdots, n
    \end{align}
    and $\gamma_j$ is the maximum information gain under the kernel $k_j$:
    \begin{align}
    \gamma_j = \mathrm{max} \dfrac{1}{2}\mathrm{log}(\mathrm{det}(I_N-\sigma_{noise}^{-2}K_j(x,x')))
    \end{align}
    where $x,x'\in\{x^{(1)}, \cdots, x^{(N)}\}.$
    }
\end{lemma}

Then we have $\mu_{gp} = [\mu_1, \cdots, \mu_n]^\intercal$ and $\sigma = [\sigma_1, \cdots, \sigma_n]^\intercal$ {to compensate $\delta(x)$ and generate confidence bounds of prediction error.} According to \textbf{Lemma} \ref{lemma:gp1}, high probability statements on the maximum {prediction} error between $\mu_{gp}$ and $\delta(x)$ can be made and used for analysis in the following section.

\subsection{Control Lyapunov Function and Control Barrier Function Based Quadratic Program}
\label{sec:CLF-CBF-QP}
In this section, we will introduce how to design $\mu_{qp}$. A quadratic programming problem is solved with several constraints to guarantee the asymptotic stability, safety, and control feasibility of the system.

\subsubsection{Stability Constraint}
A CLF is used to construct a constraint to guarantee the stability of the system. Let $P \in \mathbb{R}^{2n \times 2n}$ be the unique positive definite matrix satisfying $A^\intercal P+PA=-Q$, where $Q \in \mathbb{R}^{2n \times 2n}$ is a positive definite matrix, $A = \begin{bmatrix} 0 & I \\ -K_P & -K_D \end{bmatrix}$ in (\ref{eq:errdyn2}).

\begin{lemma}
    \label{lemma:clfcondition}
    Consider the system (\ref{eq:dynamics}) with a bounded desired state $x_{d}$. The proposed control strategy ensures that the tracking error $e$ semi-globally asymptotically converges to zero with probability at least $(1-\varsigma)^n$ for {compact} $e \in \mathcal{E}$
    %=\{e \in \mathbb{R}^{2n} \vert e^\intercal Pe \leq e(0)^\intercal Pe(0)\}$
    if:
    \begin{equation}
    \label{eq:clfcondition}
        2e^\intercal P B \mu_{qp} + \dfrac{e^\intercal Pe}{\epsilon} + 2\Vert e^\intercal PB \Vert \cdot \Vert \beta \Vert \cdot \Vert \sigma \Vert \leq 0.
    \end{equation}
\end{lemma}
\textit{Proof:} Consider a candidate Lyapunov function $V(e)=e^\intercal Pe$. We get $\dot{V}(e) = -e^\intercal Qe + 2e^\intercal PB(\mu_{qp} - \mu_{gp} + \delta)$ from (\ref{eq:errdyn2}). Let $\omega = e^\intercal PB$ for simplicity, then we have:
\begin{align}
    \dot{V}(e) + \dfrac{V(e)}{\epsilon} &= -e^\intercal Qe + 2\omega(\mu_{qp} - \mu_{gp} + \delta) + \dfrac{e^\intercal Pe}{\epsilon} \notag  \\   
    \label{eq:dotveplusve2}
    &< 2\omega\mu_{qp} + \dfrac{e^\intercal Pe}{\epsilon} + 2\omega(\delta - \mu_{gp}) \\
    \label{eq:dotveplusve3}
    &\leq 2\omega\mu_{qp} + \dfrac{e^\intercal Pe}{\epsilon} + 2\Vert \omega \Vert \cdot \Vert \delta - \mu_{gp} \Vert
\end{align}
where $\epsilon > 0$ is a positive constant. We get (\ref{eq:dotveplusve2}) as $-e^\intercal Qe < 0$ with $Q$ a positive definite matrix and the inequality (\ref{eq:dotveplusve3}) comes from the Cauchy-Schwarz inequality. From \textit{Lemma}\ref{lemma:gp1} and (\ref{eq:dotveplusve3}), we have:
$Pr\{ \dot{V}(e) + \dfrac{V(e)}{\epsilon} \leq 2\omega\mu_{qp} + \dfrac{e^\intercal Pe}{\epsilon} + 2\Vert \omega \Vert \cdot \Vert \beta \Vert \cdot \Vert \sigma \Vert , \forall e \in \mathcal{E} \} \geq (1 - \varsigma)^n$. It yields:
\begin{equation}
    Pr\{ \dot{V}(e) + \dfrac{V(e)}{\epsilon} < 0 , \forall e \in \mathcal{E} \backslash \{0\} \} \geq (1 - \varsigma)^n.
\end{equation}

Then exponentially asymptotic stability is guaranteed \cite{Ames2014rapidly} at a probability of at least $(1-\varsigma)^n$ with control input under condition (\ref{eq:clfcondition}).
Now we can construct a constraint for the quadratic programming problem as below:
\begin{align}
    \label{eq:clfconstraint}
    &H_{clf}\mu_{qp} + b_{clf} \leq d_1
\end{align}
where,
\begin{align}
    \notag
    &H_{clf}=2 \omega,  \\
    \notag
    &b_{clf}=\dfrac{e^\intercal Pe}{\epsilon} + 2 \Vert \omega \Vert \cdot \Vert \beta \Vert \cdot \Vert \sigma \Vert. 
\end{align}

The relaxation variable $d_1$ allows the quadratic programming solver to find a solution satisfying other incorporate constraints, e.g., control constraint, risking losing the convergence of the tracking error $e$ to 0. In practice, the relaxation variable $d_1$ will be optimized at each timestep and penalized by a large parameter.

\subsubsection{Safety Constraint}
We leverage control barrier functions(CBFs) \cite{Ames2014control} to derive constraints that guarantee the safety of the system. A safety set $\mathcal{S}$ is specified in which the state of the system is considered safe and is defined as the 0-superlevel set of a continuous differentiable function $h:\mathbb{R}^n\rightarrow\mathbb{R}$:
\begin{equation}
    \mathcal{S}=\{x \in \mathbb{R}^n \vert h(x) \geq 0\}.
\end{equation}
\begin{definition}\cite{Khalil2002nonlinear}
    A set $\mathcal{S}$ is forward invariant if for every $x_0 \in \mathcal{S}$, $x(t, x_0) \in \mathcal{S}$ for all $t \in \mathbb{R}_0^+$. The system (\ref{eq:dynamics}) is safe with respect to the set $\mathcal{S}$ if the set $\mathcal{S}$ is forward invariant.
\end{definition}

\begin{definition}
\label{def:cbf}
    Let $B(x)$ be a candidate control barrier function. If there exists a class-K function $\gamma$ such that $\dot{B}(x) \leq \gamma(h(x))$, then $B(x)$ is called a control barrier function.
\end{definition}
We first rewrite (\ref{eq:errdyn1}) as:
\begin{align}
    % \label{eq:xdyn1}
    \notag
    \dot{x}=\begin{bmatrix}\dot{x}_1 \\ \dot{x}_2\end{bmatrix}
    &=\begin{bmatrix} 0 & I \\ 0 & 0 \end{bmatrix}x
    +\begin{bmatrix} 0 \\ I \end{bmatrix}(\mu_{d} + \mu_{pd} + \mu_{qp} -\mu_{gp} + \delta(x)) \\
    \label{eq:xdyn2}
    &=A_0x+B_0(\mu_{d} + \mu_{pd} + \mu_{qp} -\mu_{gp} + \delta(x)).
\end{align}
\begin{lemma}
    Consider the system (\ref{eq:dynamics}) with a bounded desired state $x_{d}$. The proposed control strategy ensures the safety of the system with a probability of at least $(1-\varsigma)^n$ if:
    \begin{align}
        \notag
        \frac{\partial B}{\partial x}^{\intercal} B_0 \mu_{qp} &- \gamma (h(x)) + \frac{\partial B}{\partial x}^{\intercal} (A_0 x + B_0(\mu_{d} + \mu_{pd}))  \\
        \label{eq:cbfcondition}
        &+ \left \Vert \frac{\partial B}{\partial x}^{\intercal} B_0 \right \Vert \cdot \Vert \beta \Vert \cdot \Vert \sigma \Vert \leq 0.
    \end{align}
\end{lemma}
\textit{Proof:} {Similar to the proof of Lemma \ref{lemma:clfcondition}, with $\dot{B}(x) = \frac{\partial B}{\partial x}^{\intercal} \dot{x}$,} we have:
\begin{align}
    \notag
    \dot{B}(x) - \gamma (h(x)) = &\frac{\partial B}{\partial x}^{\intercal} (A_0x+B_0(\mu_{d} + \mu_{pd})) + \frac{\partial B}{\partial x}^{\intercal} B_0 \mu_{qp} +    \\
    \notag
    & \frac{\partial B}{\partial x}^{\intercal} B_0 (\delta(x) - \mu_{gp}) - \gamma (h(x)) \\
    \notag
    \leq &\frac{\partial B}{\partial x}^{\intercal} B_0 \mu_{qp} - \gamma (h(x)) +   \\
    \notag
    & \frac{\partial B}{\partial x}^{\intercal}(A_0x+B_0(\mu_{d} + \mu_{pd})) +   \\
    \label{eq:dotkxplusgammahx2}
    & \left \Vert \frac{\partial B}{\partial x}^{\intercal} B_0 \right \Vert \cdot \Vert \delta(x) - \mu_{gp} \Vert.
\end{align}
From \textit{Lemma}\ref{lemma:gp1} and (\ref{eq:dotkxplusgammahx2}), we have:
$Pr\{ \dot{B}(x) - \gamma (h(x)) \leq \frac{\partial B}{\partial x}^{\intercal} B_0 \mu_{qp} - \gamma (h(x)) + \frac{\partial B}{\partial x}^{\intercal}(A_0x+B_0(\mu_{d} + \mu_{pd})) + \left \Vert \frac{\partial B}{\partial x}^{\intercal} B_0 \right \Vert \cdot \Vert \beta \Vert \cdot \Vert \sigma \Vert \} \geq (1 - \varsigma)^n$. It yields:
\begin{equation}
    Pr\{ \dot{B}(x) - \gamma (h(x)) \leq 0 \} \geq (1 - \varsigma)^n.
\end{equation}

Then (\ref{eq:cbfcondition}) is a sufficient condition for safety at a probability of at least $(1-\varsigma)^n$. Now we can construct a constraint  as below:
\begin{align}
    \label{eq:cbfconstraint}
    &H_{cbf}\mu_{qp} + b_{cbf} \leq d_2
\end{align}
where,
\begin{align}
    \notag
    H_{cbf}=& \frac{\partial B}{\partial x}^{\intercal} B_0,  \\
    \notag
    b_{cbf}= & - \gamma (h(x)) + \frac{\partial B}{\partial x}^{\intercal}(A_0x+B_0(\mu_{d} + \mu_{pd}))    \\
    \notag
    & + \left \Vert \frac{\partial B}{\partial x}^{\intercal} B_0 \right \Vert \cdot \Vert \beta \Vert \cdot \Vert \sigma \Vert. 
\end{align}

Similar to (\ref{eq:clfconstraint}), a relaxation variable $d_2$ is used here. 
This relaxation variable also helps to ensure the feasibility of the quadratic program. Safety is still guarantee as long as $d_2 \leq 0$. When the violation of safety constraints is inevitable due to control constraints, the quadratic program still help avoid damage as much as possible.

\subsubsection{Quadratic Program}
Considering the control constraints, we can now construct a quadratic program to obtain $\mu_{qp}$ which guarantee asymptotic stability and safety of the system with (\ref{eq:clfconstraint}) and (\ref{eq:cbfconstraint}) as below:
\begin{align}
\label{eq:clf_cbf_qp}
    \arg\min_{\mu_{qp},d_1,d_2} \quad & \mu_{qp}^\intercal \mu_{qp} + p_1d_1^2 + p_2d_2^2,\\
    s.t. \quad & H_{clf}\mu_{qp} + b_{clf} \leq d_1, \tag{\textbf{CLF Constraint}}\\
    & H_{cbf}\mu_{qp} + b_{cbf} \leq d_2, \tag{\textbf{CBF Constraint}}\\
    & H_{u}\mu_{qp} + b_{u} \leq 0 \tag{\textbf{Control Constraint}}
\end{align}
where \begin{small}$H_u = [-g^{-1}(x), g^{-1}(x)]^\intercal,\quad b_u = [u_{min} - g^{-1}(x)(\mu_{d} + \mu_{pd} - \mu_{gp} - \hat{f}(x)), - u_{max} + g^{-1}(x)(\mu_{d} + \mu_{pd} - \mu_{gp} - \hat{f}(x))]^\intercal$\end{small}.
{As the penalty parameters $p_1$ and $p_2$ are greater than zero, the QP is a convex optimization problem with several linear inequality constraints which can be solved in (weakly) polynomial time \cite{Kozlov1980the}.}

\subsection{Model Predictive Control}
\label{sec:mpc}
Suppose we are given a reference trajectory $\mathcal{T}$ as:
\begin{equation}
    \mathcal{T} = \{ x_{ref}(t) \in X \subset \mathbb{R}^{2n} | t \in (t_0, T) \}
\end{equation}
where $x_{ref}(t) = [x_{1ref}(t), x_{2ref}(t)]^\intercal $, $t_0$ and $T$ are the initial and end time of the trajectory respectively. Based on the result of feedback linearization, we construct the MPC scheme based on the linearized system dynamics as:
\begin{equation}
    \label{eq:mpcdynamics}
    \dot{x}_{1ref}(t) = x_{2ref}(t),\quad \dot{x}_{{2}ref}(t) = \mu_{2ref}(t).
\end{equation}

An MPC algorithm is formulated in discrete time by solving an online open-loop finite-horizon optimal control problem (OCP) at each sampling time $t_k = t_0 + k \cdot dt$, where $k \in \mathbb{N}$, $dt$ is the control period. The OCP is specified as:
\begin{align}
    \label{eq:mpcocp}
    \min_{\mu_{mpc}(t)} \quad & \mathcal{J}(x(t_k), \mu_{mpc}(t_k)),   \\
    \notag
    s.t. \quad& \bar{x}_1(t)=\bar{x}_2(t),\quad \bar{x}_2(t) = \mu_{mpc}(t), \\
    \notag
    & \bar{x}(t_k) = x(t_k),  \\
    \notag
    & \bar{x}(t) \in X, \\
    \notag
    & \mu_{mpc}(t) \in \mathcal{U}
\end{align}
where $\bar{x}(t) = [\bar{x}_1(t), \bar{x}_2(t)]^\intercal$ and $\mu_{mpc}$ are the predicted state and control respectively.
The MPC objective function $\mathcal{J}$ is constructed as:
\begin{align}
    \label{eq:mpcobj}
    \notag
    &\mathcal{J}(x(t_k), \mu_{mpc}(t_k)) \\
    = &\sum_{t_k}^{t_k+K \cdot dt} \Vert \bar{x}(t) - x_{ref}(t) \Vert_{Q_{mpc}}^2 + \Vert \mu_{mpc}(t) \Vert_{R_{mpc}}^2
\end{align}
where $K$ is the predictive step, positive semi-definite matrix $Q_{mpc} \in \mathbb{R}^{2n \times 2n}$ weights the tracking error between the predictive states and reference states, positive definite matrix $R_{mpc} \in \mathbb{R}^{n \times n}$ ensures regularization of the inputs. Based on the linearized system (\ref{eq:mpcdynamics}), the OCP is a convex quadratic program that can be solved quickly in real-time with efficient methods.

At each sampling time $t_k$, the {MPC} takes current measured state and reference trajectory as input and is solved to obtain the optimal states $\bar{x}(t)$ and control $\mu_{mpc}(t), t=t_k + i \cdot dt, i = 0, \cdots , K-1$. The first step of the state and control are applied as $x_d = \bar{x}(t_k)$ and $\mu_{d} = \mu_{mpc}(t_k)$ respectively.

\subsection{Event-triggered Model Update}
\label{sec:et}
In the proposed method, {the} model error $\delta$ is learned and predicted by online GPs, in which model update {is important for model fidelity}. As is common in GPs, the measurements are not only used to update the dataset $\mathcal{D}$ in (\ref{eq:gpdataset}), but also used in setting the approximation properties for the covariance function. We employed the maximization of the log marginal likelihood to update the hyperparameters $\sigma_f$ and $l$ of the covariance function (\ref{eq:gpkernel}):%, which realize the adaption of the GPs:
\begin{align}
    \label{eq:logmarli}
    \log \  p(\mathbf{y}|x^*, \mathbf{\theta}) = - \frac{1}{2} \mathbf{y}^\intercal \mathbf{K}_y^{-1} \mathbf{y} - \frac{1}{2} \log |\mathbf{K_y}| - \frac{N}{2} \log 2\pi,  \\
    \frac{\partial}{\partial \theta_i} \log \  p(\mathbf{y}|x^*, \mathbf{\theta}) = - \frac{1}{2} \mathbf{y}^\intercal \mathbf{K}^{-1} \frac{\partial \mathbf{K}}{\partial \theta_i} \mathbf{K}^{-1} \mathbf{y} + \frac{1}{2}\mathrm{tr}(K_j^{-1} \frac{\partial \mathbf{K}}{\partial \theta_i}).
\end{align}
where $\theta$ are hyperparameters and \begin{small}$\mathbf{K_y} = K_j + \sigma_{noise}^2I$ \end{small} in (\ref{eq:gpK}). The maximization of (\ref{eq:logmarli}) is obtained by solving a optimization problem with a non-convex objective function, which brings much computational burden.

Thus, it is reasonable to update the GPs asynchronously (whenever needed) but not synchronously. As is mentioned in \ref{sec:CLF-CBF-QP}, asymptotic stability of the system are guaranteed under condition (\ref{eq:clfcondition}). However, it is predictable that when {a prediction with high uncertainty (which is reflected in the predict variance in (\ref{eq:gpvar})) is made}, (\ref{eq:clfcondition}) will be hard to be satisfied. The quadratic program are still solvable {with the help of} the relaxation variable $d_1$ and $d_2$, but stability and safety are hardly guaranteed with such an imprecise prediction {of} GPs. It is reasonable to update the GPs in such a situation to improve the prediction performance of GPs and {maintain the fidelity of models so that the stability constraints can be satisfied.} As a result, we propose the following event-triggering condition:
\begin{equation}
    \label{eq:trigger}
    2e^\intercal P B \mu_{qp} + \dfrac{e^\intercal Pe}{\epsilon} + 2\Vert e^\intercal PB \Vert \cdot \Vert \beta \Vert \cdot \Vert \sigma \Vert > 0
\end{equation}
where $\mu_{qp}$ is the result of the quadratic program (\ref{eq:clf_cbf_qp}). Whenever the condition (\ref{eq:trigger}) is satisfied, we update the GPs to obtain a precise prediction with which stability and safety of the system are guaranteed.

\section{Simulations}
\label{sec:simulation}
In this section, the proposed approach is applied to a quadrotor model for trajectory tracking tasks under uncertain disturbances.
The effectiveness of the proposed approach is validated via simulations.

\subsection{Dynamics and Control}
The states of the quadrotor can be modeled as
\begin{align}
    \dot{p} &= v,    \\
    \dot{v} &= g + \frac{1}{m}Rf_u+\delta
\end{align}
where the position and velocity are described by $p\in \mathbb{R}^3$ and $v\in \mathbb{R}^3$, $g$ is gravity, $m$ is the mass of quadrotor, $f_u$ is the quadrotor thrust, and $\delta$ is disturbances.
The attitude rotation matrix $R$ from body frame to global frame can be written as
\begin{equation}
    R=\begin{bmatrix}
        c\theta c\psi  &  s\phi s\theta c\psi-c\phi s\psi  &  c\phi s\theta c\psi+s\phi s\psi \\
        c\theta s\psi  &  s\phi s\theta s\psi+c\phi c\psi  &  c\phi s\theta c\psi-s\phi c\psi \\
        -s\theta       &  s\phi c\theta                    &  c\phi c\theta
    \end{bmatrix}
\end{equation}
{where $\phi, \theta, \psi$ are roll, pitch and yaw respectively, and $c, s$ refer to $cos$ and $sin$.}
To fit the form in (\ref{eq:dynamics}), we transform the dynamics model as
\begin{align}
    \notag
    x &= [x_1,\ x_2]^\intercal = [p,\ v]^\intercal \\
      &= [p_x,\ p_y,\ p_z,\ v_x,\ v_y,\ v_z]^\intercal
\end{align}
with thrust force $f_u$ as controls.

\begin{figure*}[t]
\centering
\subfigure[Trajectory]{
\begin{minipage}[t]{0.33\linewidth}
\centering
\includegraphics[width=1.9in]{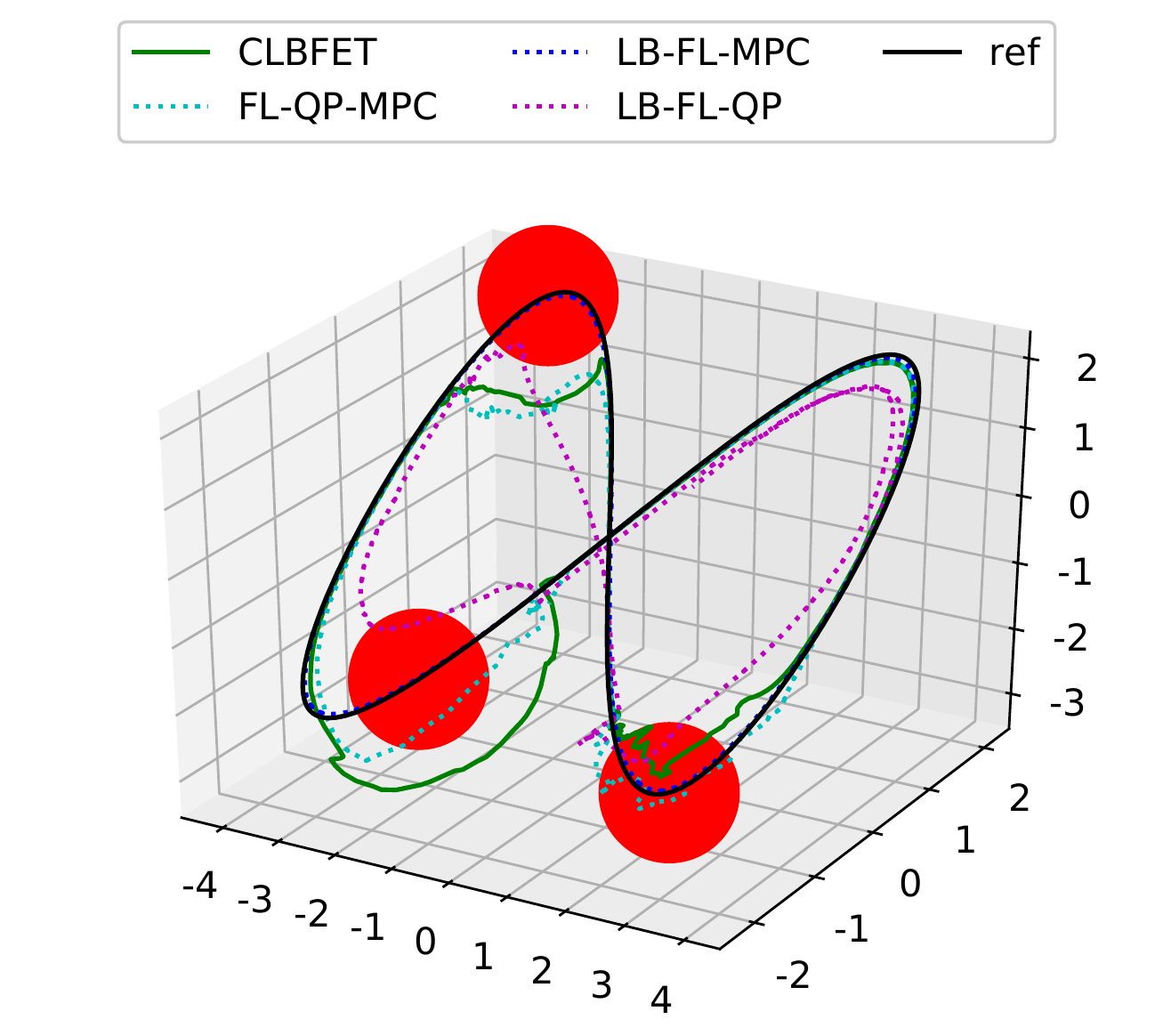}
\end{minipage}%
\label{fig:A1}
}%
\subfigure[Tracking error]{
\begin{minipage}[t]{0.33\linewidth}
\centering
\includegraphics[width=2.2in]{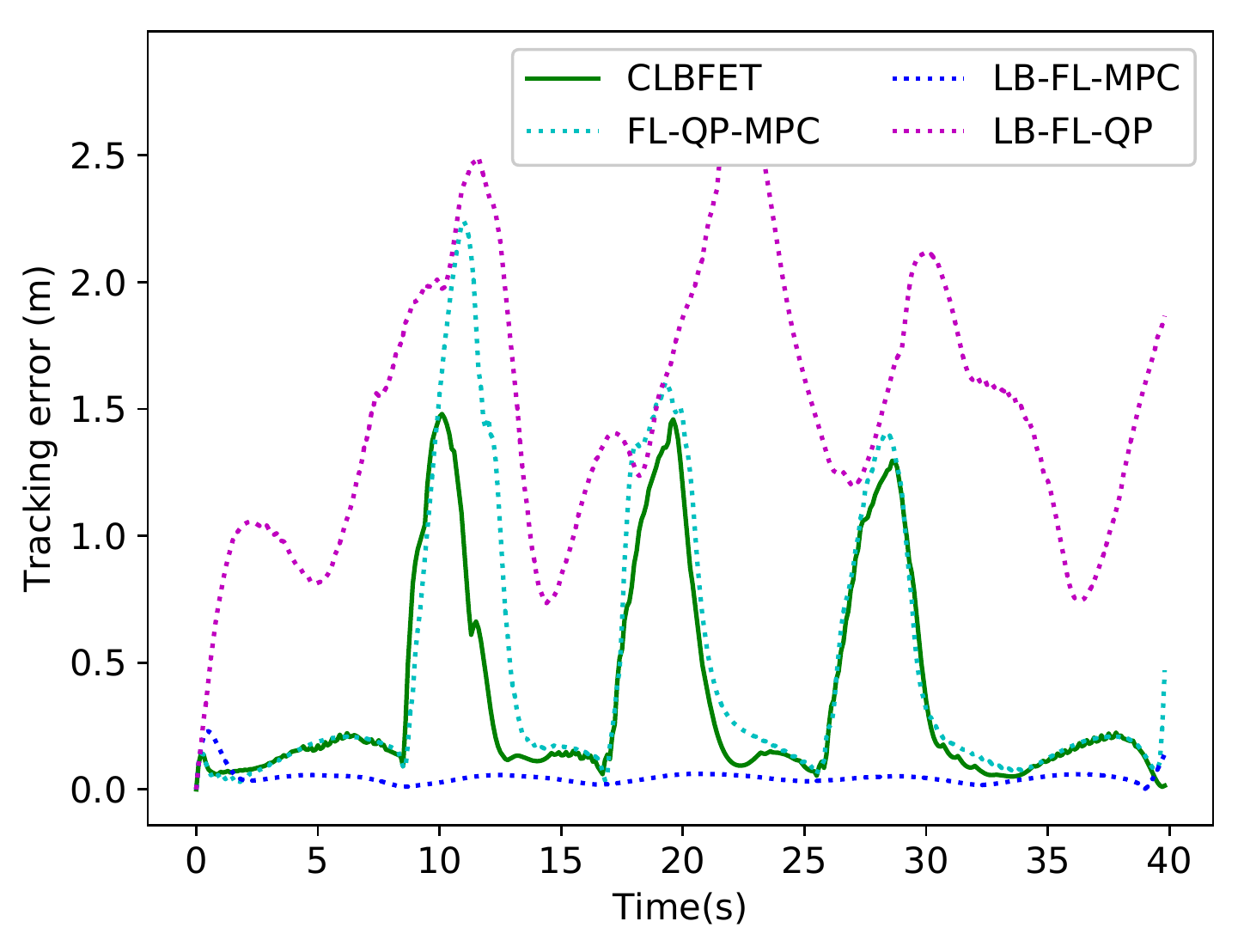}
\end{minipage}%
\label{fig:A3}
}%
\subfigure[Distance to obstacles]{
\begin{minipage}[t]{0.33\linewidth}
\centering
\includegraphics[width=2.1in]{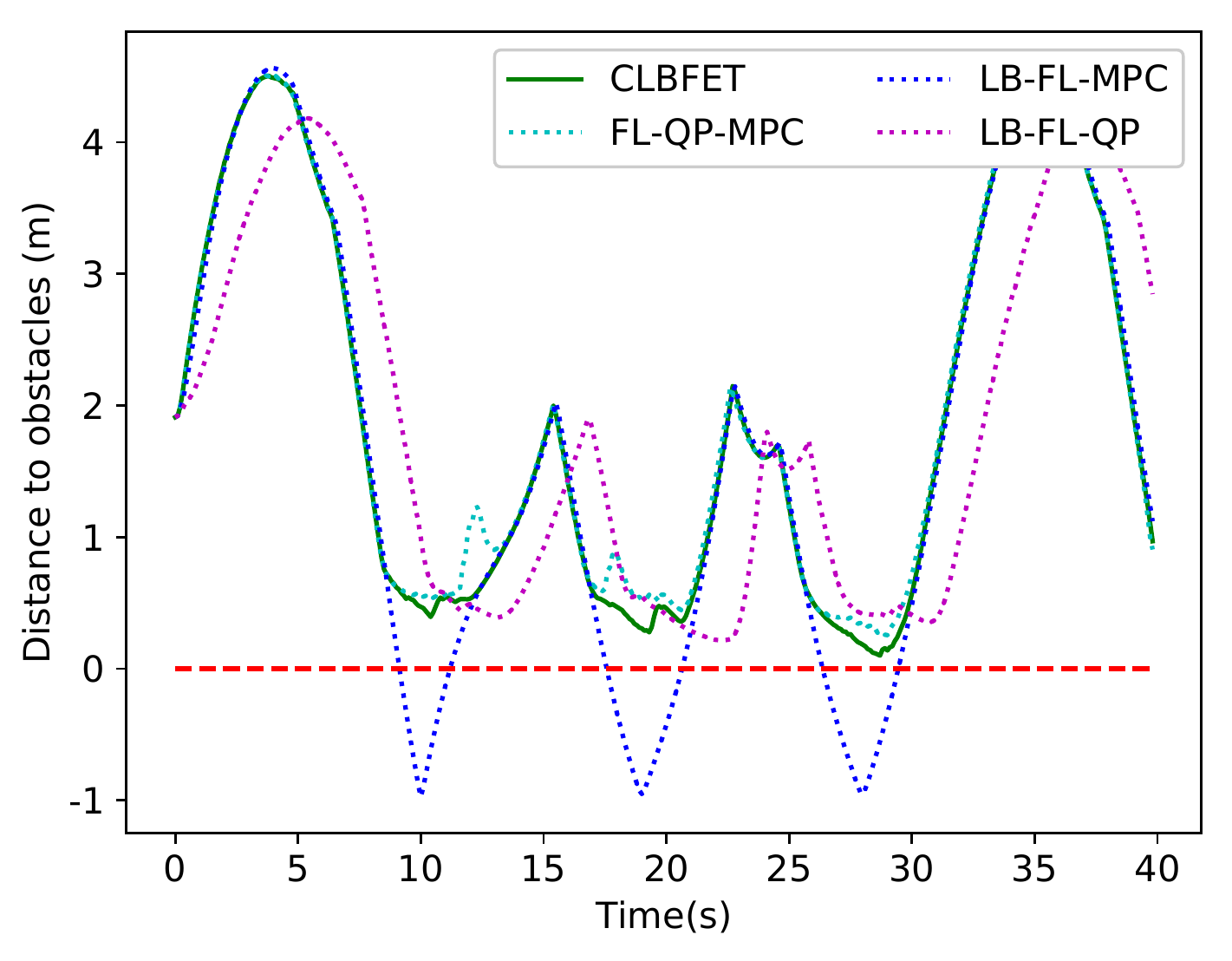}
\end{minipage}
\label{fig:A5}
}%
\centering
\caption{(a) The tracking performance of the four methods in the ablation experiment. The red spheres are the three obstacles. ref is the reference trajectory. (b) The tracking errors to the reference trajectory. (c) The distances to the nearest obstacles.}
\end{figure*}

\subsection{Simulation Setup}
A simulation platform is used on an Intel Xeon X5675 CPU with 3.07 GHz clock frequency in Python 3.6 code which has not been optimized for speed.
One iteration of the {proposed} algorithm for this problem takes less than 10ms on {the} platform.
The control gains matrix in PD controller is $K_p = {diag[1, 1, 1]}$ and $K_d = {diag[1, 1, 1]}$ in (\ref{eq:qpcontrol}).
{The parameters of PD controller gains are set the same in BALSA\cite{Fan2020bayesian} in the comparative experiment, and are not specifically optimized in BALSA and the proposed CLBFET.
ALPaCA\cite{Harrison2018meta} is used as the Bayesian modeling algorithm in BALSA.
}
We use the GPy package to build 3 GPs {for 3 dimensions}.
Hyperparameters $\sigma_f$ and $l$ in (\ref{eq:gpkernel}) will be optimized soon after the simulation begin.
We set $N=60$ for GPs and $\beta=1$ in (\ref{eq:clfcondition}) and (\ref{eq:cbfcondition}).
The MPC is constructed as a QP which is solved using OSQP solver \cite{Stellato2020OSQP}.
%The nonlinear MPC which is used in the comparative experiment is solved using CasAdi\cite{Andersson2019CasADi} with the IPOPT nonlinear programming solver\cite{Biegler2009large}.
We set $K=20$, $Q_{mpc}={diag[10,10,10,0.5,0.5,0.5]}$, $R_{mpc}={diag[0.5,0.5,0.5]}$ in (\ref{eq:mpcobj}).
The QP in (\ref{eq:clf_cbf_qp}) is also solved with the OSQP solver with {penalty coefficients} $p_1=1e8$ and $p_2=1e12$.
{The complexity of solving the QP in (\ref{eq:clf_cbf_qp}) with OSQP is $\mathcal{O}(n^3+n^2N)$, where $N$ is the number of iterations.
The number of iterations $N$ is between 25 and 100 in practice.
Time to solve MPC and CLF-CBF-QP in (\ref{eq:clf_cbf_qp}) are both less than $1ms$.}
The $Q$ matrix in the Lyapunov function is set as $Q={diag[1,1,1,1,1,1]}$ and $\epsilon=100$ in (\ref{eq:clfcondition}).
Let our safety set be $\mathcal{S} = \{x \in \mathbb{R}^n | h(x,x_{obs}) \geq 0\}$ where $x_{obs} = [p_{xobs}, p_{yobs}]^\intercal$ is the position of an obstacle and $h(x,x_{obs})$ is set as

\begin{align}
    \notag
    h(x,x_{obs}) = &\gamma_p (d - r) + \frac{v_x}{d}(p_x - p_{xobs}) + \\ 
    &\frac{v_y}{d}(p_y - p_{yobs}) + \frac{v_z}{d}(p_z - p_{zobs})
\end{align}
where $d = \left((p_x - p_{xobs})^2 + (p_y - p_{yobs})^2 + (p_z - p_{zobs})^2\right)^\frac{1}{2}$, $r$ is the radius of the obstacle and $\gamma_p=5$ in practice.
Then the barrier function $B(x,x_{obs})$ is constructed as $B(x,x_{obs})=\frac{1}{h(x,x_{obs})}$ with $\gamma = 0.08$ in Definition \ref{def:cbf}.

A lemniscate is set as a reference trajectory in the ablation experiments as shown in Fig. \ref{fig:A1}.
Five common trajectories are tested in the comparative experiments, including conical spiral, lemniscate, line, circle, and cylindrical helix, as shown in Fig. \ref{fig:B1}.
Three spherical obstacles with radius $r_{obs}=1$ are set on each trajectory at $t=10$, $t=19$ and $t=28$.

A simulated wind disturbances in \cite{Cole2018reactive} are set to numerically validate the performance of the proposed method. The disturbances $\delta(x)$ consists of three component,
\begin{equation}
    \delta(x) = \delta_c(x) + \delta_t(x) + \delta_g(x)
\end{equation}
where $\delta_c(x)$ is a constant component, $\delta_t(x)$ is a turbulent component and $\delta_g(x)$ is a gust component.
The constant component $\delta_c(x)$ is set from $3 m/s$ to $10 m/s$ randomly.
The von Kármán velocity model with low-altitude model parameters in \cite{Moorhouse1980us} is utilized to construct the turbulent component $\delta_t(x)$ as in \cite{Shinozuka1972monte} \cite{George1996simulation}.
The gust component is set as a $1-cos$ model.
We randomly generate 5 sets of parameters of the wind model in the simulations.

% \vspace{5mm}
\begin{table}[b]
\caption{Trajectory Tracking Performance in Ablation Experiment}
\label{tab:S1-1}
\begin{tabular}{c|cccc}
\hline
\multirow{3}{*}{Method} & Average      & Average     & Tracking & Collide \\
                        & Control Time & update Time & Error    & or not  \\
                        & {[}ms{]}     & {[}ms{]}    & {[}m{]}  &         \\ \hline
CLBFET                  & 8.121        & 18.48       & 0.2705   & no      \\
FL-QP-MPC             & 6.482        & -           & 0.3503   & no      \\
LB-FL-MPC             & 4.047        & 8.069       & -        & yes     \\
LB-FL-QP              & 6.094        & 19.27       & 1.273    & no      \\ \hline
\end{tabular}
\end{table}

\begin{figure*}[t]
\label{fig:B}
\centering
\subfigure[Trajectory]{
\begin{minipage}[t]{0.33\linewidth}
\centering
\includegraphics[width=1.9in]{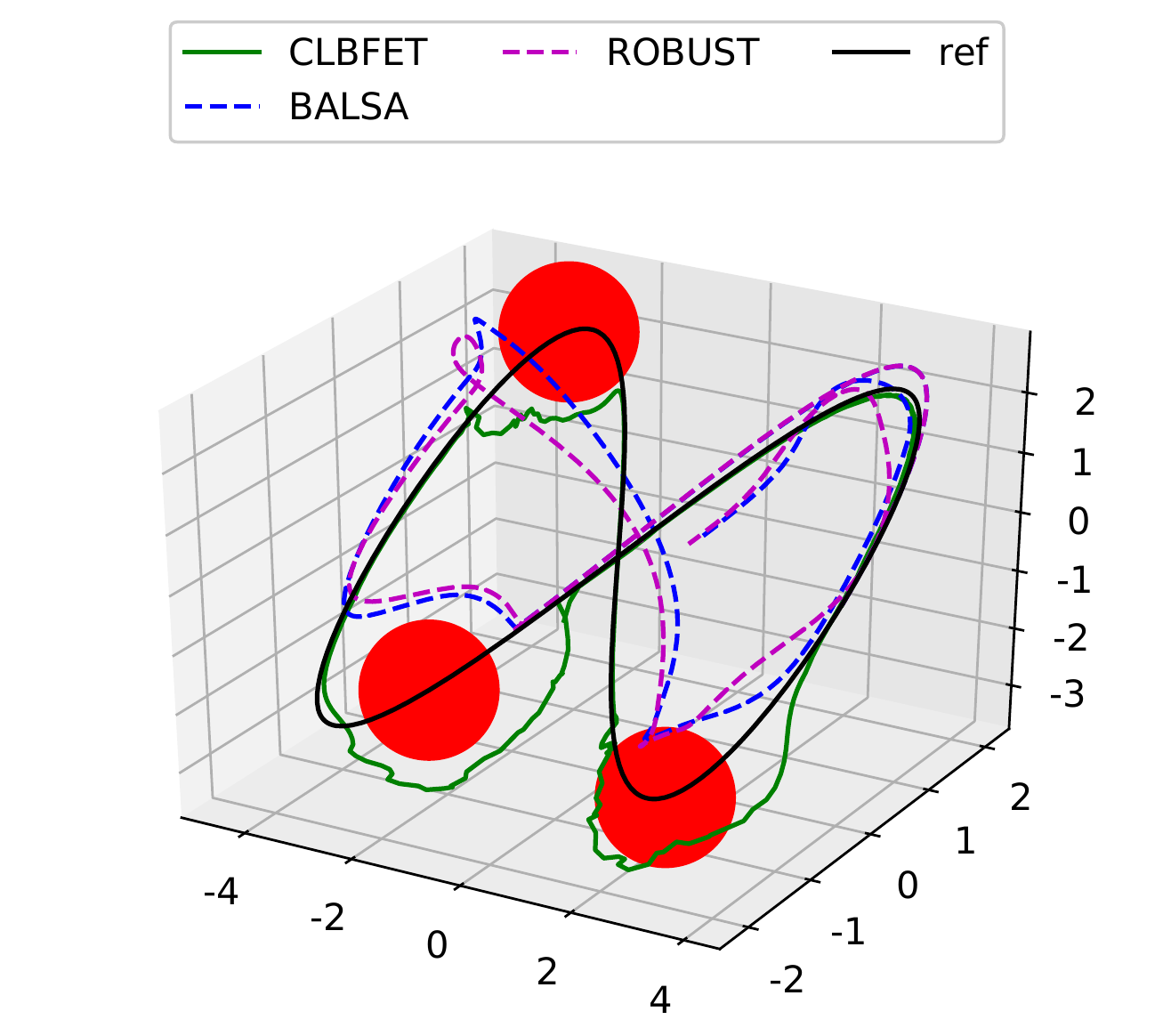}
\end{minipage}%
\label{fig:B2}
}%
\subfigure[Tracking error]{
\begin{minipage}[t]{0.33\linewidth}
\centering
\includegraphics[width=2.2in]{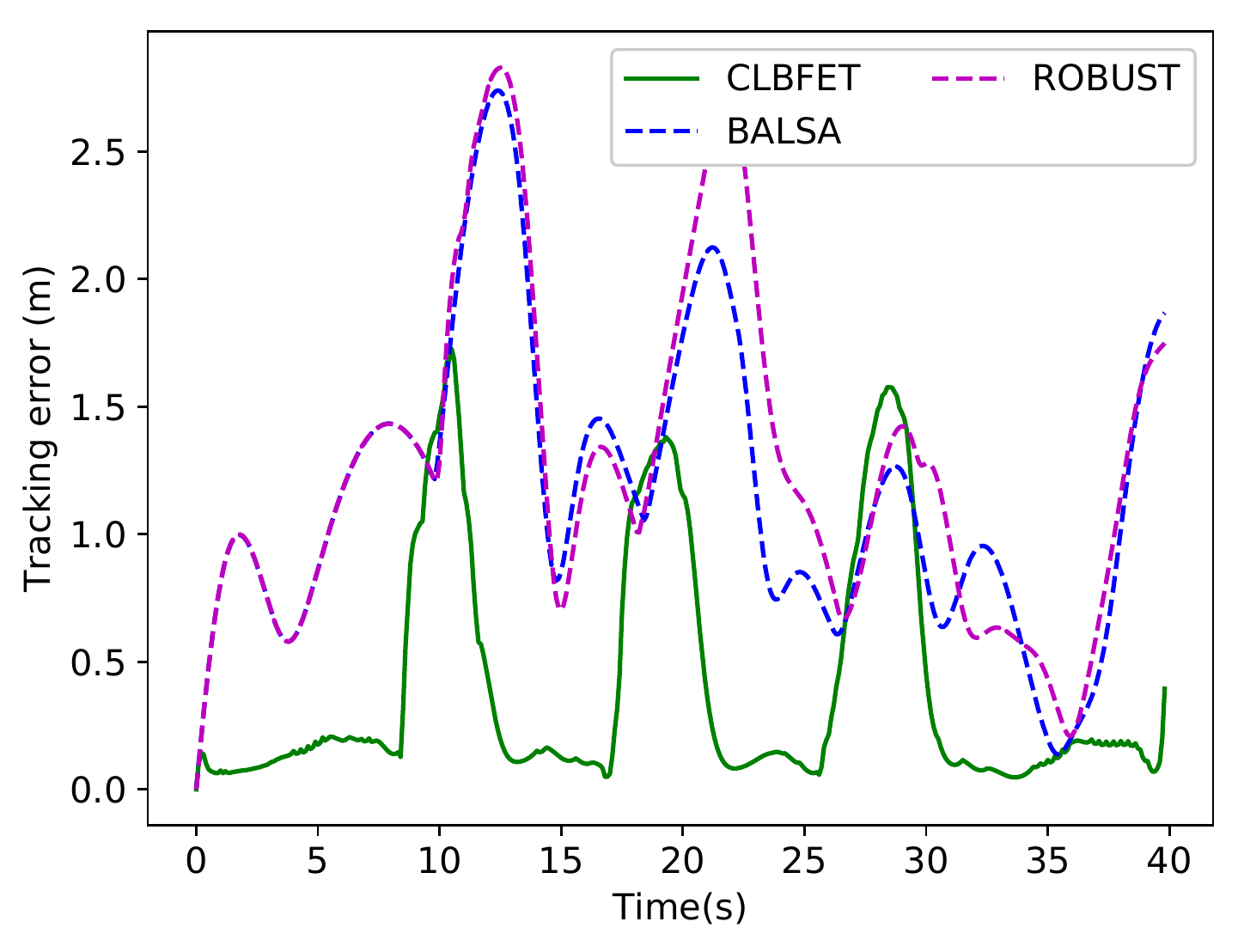}
\end{minipage}%
\label{fig:B3}
}%
\subfigure[Distance to obstacles]{
\begin{minipage}[t]{0.33\linewidth}
\centering
\includegraphics[width=2.1in]{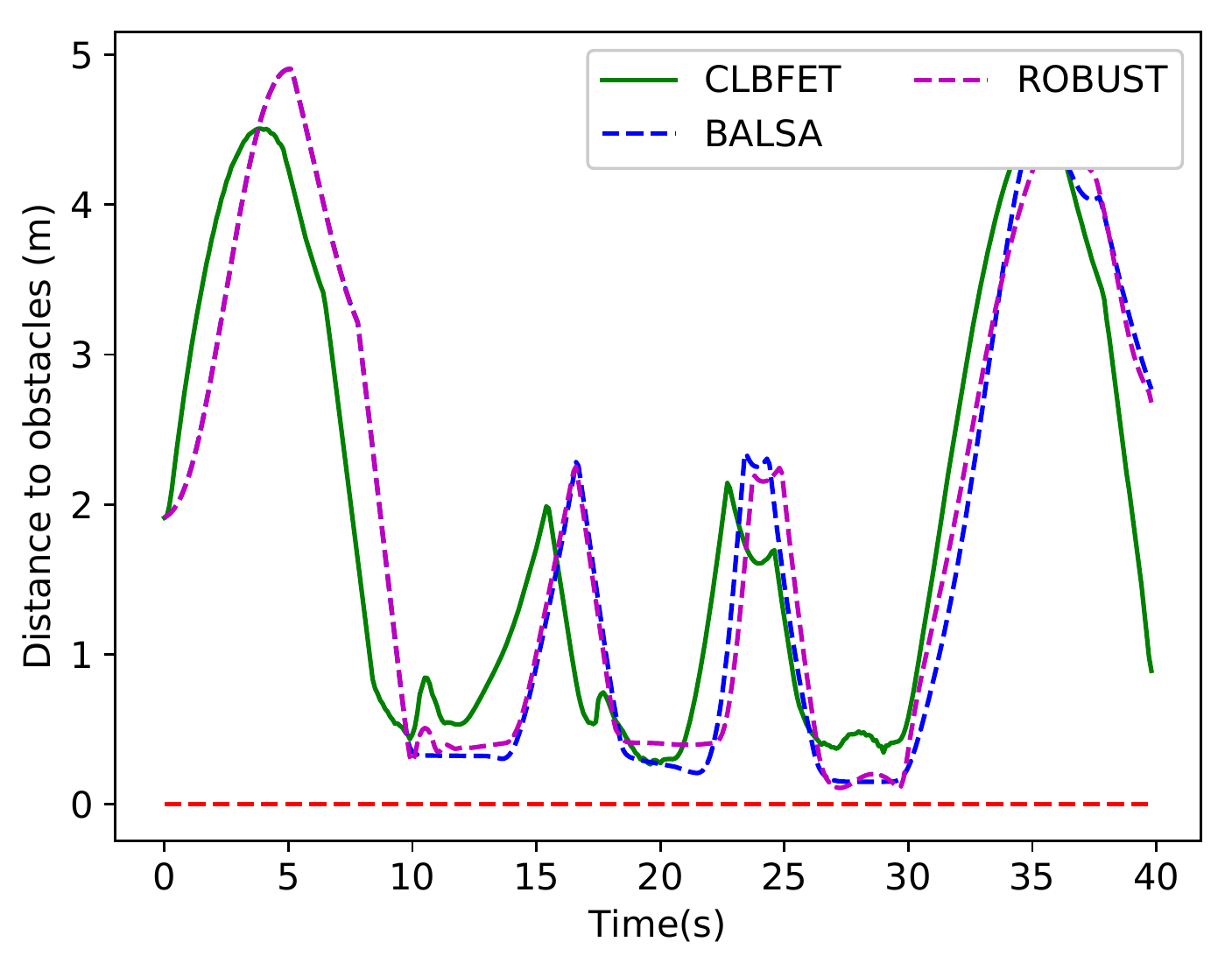}
\end{minipage}
\label{fig:B4}
}%
\centering
\caption{(a) The tracking performance in the comparative experiment. The red spheres are the three obstacles. ref is the reference trajectory. (b) The tracking errors to the reference trajectory. (c) The distances to the nearest obstacles.}
\end{figure*}

\begin{table*}[t]
\centering
\caption{Performance of different methods in different trajectories.}
\label{tab:B1}
\begin{tabular}{cc|ccccc}
\hline
\multirow{3}{*}{Trajectory} & \multirow{3}{*}{Method} & Average          & Average          & Average         & Average Distance & Minimum Distance \\
                            &                         & Control Time     & Update Time      & Tracking Error  & From Obstacles   & From Obstacles   \\
                            &                         & [ms] & [ms] & [m]      & [m]       & [m]       \\ \hline
Circle                      & CLBFET                  & 8.072            & \textbf{14.18}   & \textbf{0.7752} & \textbf{1.575}   & \textbf{1.220}   \\
Circle                      & BALSA\cite{Fan2020bayesian}                   & \textbf{4.475}   & 37.02            & 1.584           & 1.470            & 1.128            \\
Circle                      & ROBUST\cite{Nguyen2016optimal}                  & 6.165            & 36.82            & 1.543           & 1.572            & 1.072            \\ \hline
Conical Spiral              & CLBFET                  & 7.614            & \textbf{20.66}   & \textbf{0.6408} & \textbf{6.167}   & \textbf{1.404}   \\
Conical Spiral              & BALSA                   & \textbf{4.245}   & 38.25            & 2.820           & 6.065            & 1.205            \\
Conical Spiral              & ROBUST                  & 5.806            & 38.21            & 3.018           & 6.094            & 1.326            \\ \hline
Cylindrical Helix           & CLBFET                  & 8.402            & \textbf{25.35}   & \textbf{0.3572} & 2.541            & \textbf{1.224}   \\
Cylindrical Helix           & BALSA                   & \textbf{4.678}   & 36.42            & 1.196           & \textbf{2.598}   & 1.176            \\
Cylindrical Helix           & ROBUST                  & 6.543            & 36.28            & 1.300           & 2.522            & 1.077            \\ \hline
Lemniscate                  & CLBFET                  & 7.921            & \textbf{20.68}   & \textbf{0.4300} & 2.933            & \textbf{1.265}   \\
Lemniscate                  & BALSA                   & \textbf{4.352}   & 38.30            & 1.154           & 2.957            & 1.142            \\
Lemniscate                  & ROBUST                  & 5.998            & 38.43            & 1.229           & \textbf{2.999}   & 1.127            \\ \hline
Line                        & CLBFET                  & 7.614            & \textbf{4.113}   & \textbf{0.5616} & \textbf{10.90}   & 1.410            \\
Line                        & BALSA                   & \textbf{4.256}   & 38.38            & 2.752           & 10.60            & 1.314            \\
Line                        & ROBUST                  & 5.885            & 38.50            & 2.819           & 10.50            & \textbf{1.432}   \\ \hline
\end{tabular}
\end{table*}

\subsection{Numerical Results}

We first design {the} ablation experiments to illustrate the effectiveness of the proposed method.
Here we compare our method (CLBFET) with three methods.
Fig. \ref{fig:A1} and Table \ref{tab:S1-1} shows the tracking performance of the four methods.
{The abbreviation FL refers to feedback linearization control and LB refers to learning-based.}
The quadrotor using {the proposed CLBFET controller shows the lowest tracking errors}.
The FL-QP-MPC controller without GPs to estimate the disturbances exhibits higher tracking error than CLBFET shown in Fig. \ref{fig:A1} and Fig. \ref{fig:A3}.
The LB-FL-MPC controller without CLF-CBF-QP is not able to guarantee stability and safety{, which leads to collisions.}
Fig. \ref{fig:A5} shows that only {\noindent the} quadrotor using LB-FL-MPC controller collides with the obstacles.
The LB-FL-QP controller without MPC shows the worst tracking performance, which verifies the effectiveness of MPC based on linearized dynamics.
In addition, controllers with MPC make the quadrotor back to the reference trajectory faster after avoiding the obstacles as shown in Fig. \ref{fig:A3}.

{The average control and model update time are shown in Table \ref{tab:S1-1}.
The complete CLBFET controller takes the most time in control, but still reaches a satisfactory real-time performance with a control frequency of more than 100HZ.
The LB-FL-QP controller and the CLBFET show similar update times.
As a result of the lack of CLF-CBF-QP, a time-trigger instead of an event-trigger is used in the LB-FL-MPC controller, which leads to different average update time as shown in Table \ref{tab:B1}
Note that the LB-FL-MPC controller is unable to guide the quadrotor to avoid collisions without CLF-CBF-QP even if the update frequency is increased as much as possible.
}

\begin{figure}[h]
\centering
\includegraphics[scale=0.52]{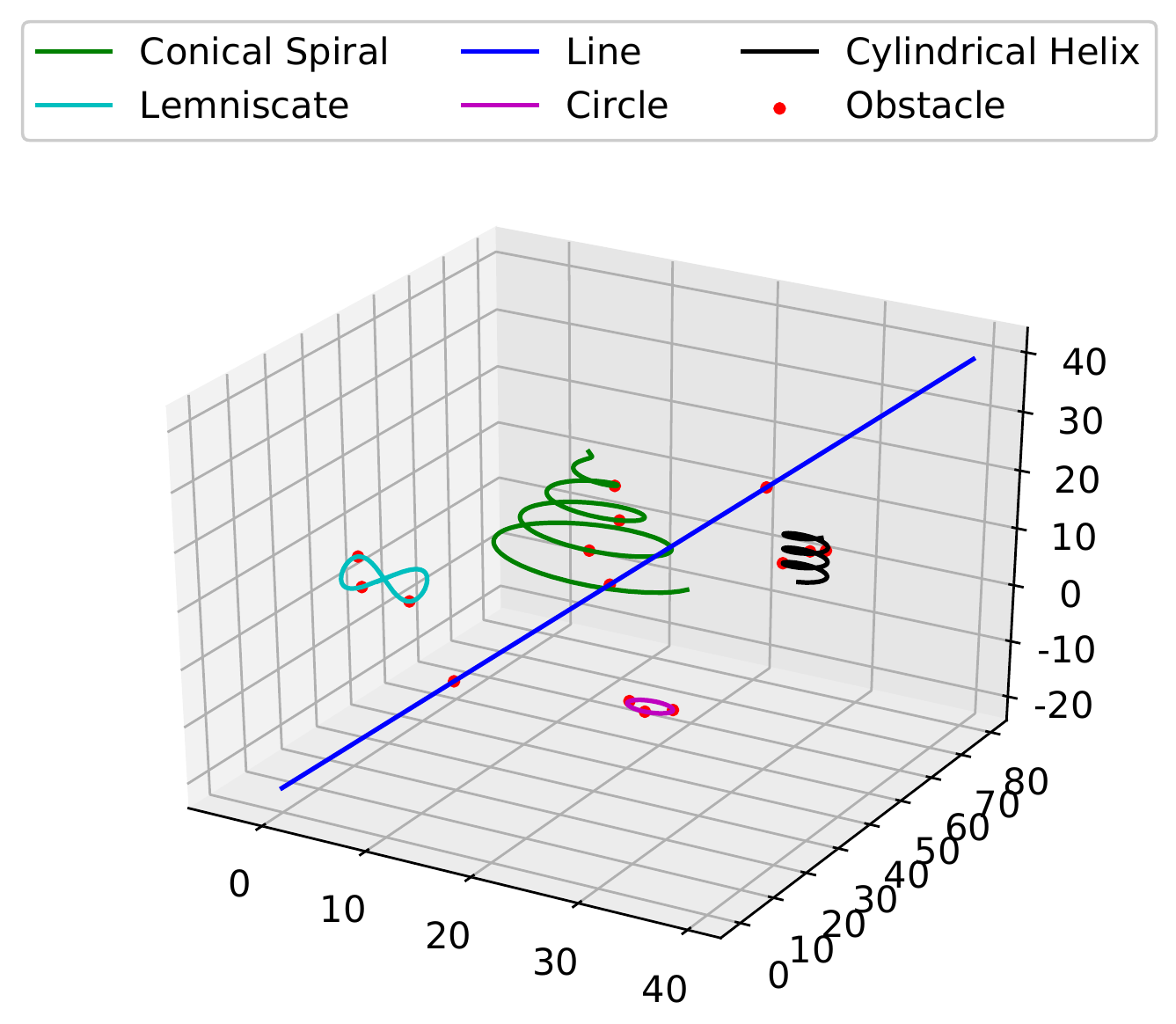}
\caption{Five trajectories set in the comparative experiment. Red points are the center of three obstacles in each trajectory.}
\label{fig:B1}
\end{figure}

In the next experiment, we compare the proposed method with BALSA \cite{Fan2020bayesian} and ROBUST \cite{Nguyen2016optimal}.
BALSA is applied to car-like vehicles in \cite{Fan2020bayesian} and adjusted to quadrotors here.
For method ROBUST, a robust-CLF-CBF-QP is used in the experiment.
Several reference trajectories are set as shown in Fig. \ref{fig:B1}.
% TODO
Fig. 3 shows the results of trajectory lemniscate.
The proposed CLBFET keeps the furthest away from the obstacles while performing best at trajectory tracking as shown in Fig. \ref{fig:B2} and Fig. \ref{fig:B3}.
The results of the comparative experiment are listed in Table \ref{tab:B1}.
{Firstly, CLBFET has satisfactory real-time performance.}
As shown in Table \ref{tab:B1}, our CLBFET cost most time in control, but can still reach a control frequency of more than 100HZ.
On the contrary, the proposed CLBFET has the least update time on average.
The update time of BALSA which uses time-triggered updates is similar in different trajectories.
However, {with the help of} the event-triggered scheme, CLBFET costs much less time in some relatively simple trajectories like line and circle.
In the complex trajectories, CLBFET still costs less time than the other methods.
{Secondly, the proposed CLBFET significantly outperforms the other two methods with respect to the tracking accuracy in all 5 trajectories with satisfactory real-time performance.
Besides, the safety constraints are well maintained by all the methods, with similar distances from obstacles.}

\section{Conclusions}
\label{sec:conclusions}
In this paper, we have designed a novel {tracking} control scheme for nonlinear systems under uncertainties and guarantee a high probability of {stability and }safety of the systems.
The theoretical analysis proves the asymptotic stability and safety of the system under high probability.
Numerical simulations show that, with the proposed CLBFET method, a quadrotor can accurately track the reference trajectories and avoid the obstacles under uncertainties.
The effectiveness of the event-triggered scheme is also validated in simulations under different trajectories.% compared with time-trigger methods.
In future work, we will {validate the proposed method in real-world} experiments.
{The scalability of the proposed method and the uncertainties in control input will also be considered.}

\balance

\nocite{*}
\bibliographystyle{IEEEtran}
\bibliography{mybib}

\end{document}